\documentclass[11pt]{article} 
\usepackage{rldmsubmit,palatino}
\usepackage{wrapfig,graphicx}

\title{The Quest for a Common Model of the Intelligent Decision Maker}

\author{
Richard S. Sutton
\\
DeepMind and the University of Alberta\\
Edmonton, Alberta, Canada  \\
\texttt{rich@richsutton.com} 
}

\begin{document}

\maketitle

\begin{abstract}
The premise of the \emph{Multi-disciplinary Conference on Reinforcement Learning and Decision Making} is that multiple disciplines share an interest in goal-directed decision making over time. The idea of this paper is to sharpen and deepen this premise by proposing a perspective on the decision maker that is substantive and widely held across psychology, artificial intelligence, economics, control theory, and neuroscience, which I call the \emph{common model of the intelligent agent}. The common model does not include anything specific to any organism, world, or application domain. The common model does include aspects of the decision maker's interaction with its world (there must be input and output, and a goal) and internal components of the decision maker (for perception, decision-making, internal evaluation, and a world model). I identify these aspects and components, note that they are given different names in different disciplines but refer essentially to the same ideas, and discuss the challenges and benefits of devising a neutral terminology that can be used across disciplines. It is time to recognize and build on the convergence of multiple diverse disciplines on a substantive common model of the intelligent agent.
\end{abstract}


\acknowledgements{
The author gratefully acknowledges that his perspectives on these topics have been immeasurably improved by discussions with Joseph Modayil, Adam White, Marlos Machado, Andy Barto, Satinder Singh, David Silver, Doina Precup, Ben Van Roy, Patrick Pilarski, Alex Kearney, Elliot Ludvig, and James Kehoe. For any remaining na\"ivete or hyperbole, the author alone is responsible. This work was supported by funding from the Alberta Machine Intelligence Institute (Amii), DeepMind, and CIFAR.\hfill\break
\newline
\newline
\vspace*{2.5in}
\hrule
\noindent
This paper will appear as an extended abstract at the fifth \emph{Multi-disciplinary Conference on Reinforcement Learning and Decision Making}, held in Providence, Rhode Island, on June 8-11, 2022.
}

\startmain 

\section{The Quest}

The premise of the Multi-disciplinary Conference on Reinforcement Learning and Decision Making (RLDM) is that there is value in all the disciplines interested in ``learning and decision making over time to achieve a goal" coming together and sharing perspectives. The natural sciences of psychology, neuroscience, and ethology, the engineering sciences of artificial intelligence, optimal control theory, and operations research, and the social sciences of economics and anthropology---all focus in part on intelligent decision makers.
The perspectives of the various disciplines are different, but they have common elements. One cross-disciplinary goal is to identify the common core---those aspects of the decision maker that are common to all or many of disciplines. To the extent that such a common model of the decision maker can be established, the exchange of ideas and results may be facilitated, progress may be faster, and the understanding gained may be more fundamental and longer lasting. 

The quest for a common model of the decision maker is not new. 
One measure of its current vitality is the success of cross-disciplinary meetings such as RLDM and the Conference on Neural Information Processing Systems (NeurIPS), and journals such as Neural Computation, Biological Cybernetics, and Adaptive Behavior. There have been many scientific insights gained from cross-disciplinary interactions, such as the now-widespread use of Bayesian methods in psychology, the reward-prediction error interpretation of dopamine in neuroscience (Schultz, Dayan \& Montague 1997), 
and the long-standing use of the neural-network metaphor in machine learning.
Although significant relationships between many of these disciplines are as old the disciplines themselves, they are still far from settled. To find commonalities among disciplines, or really even within one discipline, one must overlook many disagreements. We must be selective. We must look for the big picture without expectation that there will be no exceptions. 

In this short paper I hope to advance the quest for a model of the intelligent decision-maker that resonates across disciplines in the following small ways. First, I explicitly identify the quest as distinct from fruitful cross-disciplinary interaction. Second, I highlight the formulation of goals as the maximization of a cumulative numerical signal as highly inter-disciplinary. Third, I highlight a particular internal structure of the decision maker---as four principal components interacting in a specific way---as already being common to multiple disciplines. Finally, I highlight terminological differences that obscure the commonalities between fields and offer terms that instead encourage the multi-disciplinary mindset.

\section{Interface Terminology}

The decision-maker makes its decisions over time, which may be divided into discrete steps at each of which new information is received and a decision is made that may affect the information that is received later. That is, there is an interaction over time with signals exchanged. What terminology shall we use for the signals and for the entities exchanging them? In psychology, the decision maker is the ``organism," receiving ``stimuli" and sending ``responses" to its ``environment." In control theory, the decision-maker is termed the ``controller," receiving ``state" and sending ``control signals" to the ``plant." Other fields use still other terms, but these illustrate the challenge---to find terms that do not prejiduce the reader towards one field or the other, but rather facilitate thinking that crosses disciplinary boundaries.

A good way to start in establishing terminology is to clarify the ideas that the words are meant to convey---and not convey. The latter is particularly significant for us, as we do not want our terms to evoke intuitions that are specific to any particular discipline. For example, calling the decision-maker an ``organism" would interfere with thinking of it as a machine, as we would in artificial intelligence. The essence of a decision maker is that it acts with some autonomy, is sensitive to its input, and has an purposeful effect on its future input. A good word for this is \emph{agent}, for which my dictionary offers the following definition: ``a person or thing that takes an active role or produces a specified effect." The word is commonly used this way within artificial intelligence for decision-makers that could be either machines or people. The term ``agent" is also preferable to "decision maker" because it connotes autonomy and purposiveness.

\begin{wrapfigure}{R}{0.30\textwidth}
\includegraphics[width=0.30\textwidth]{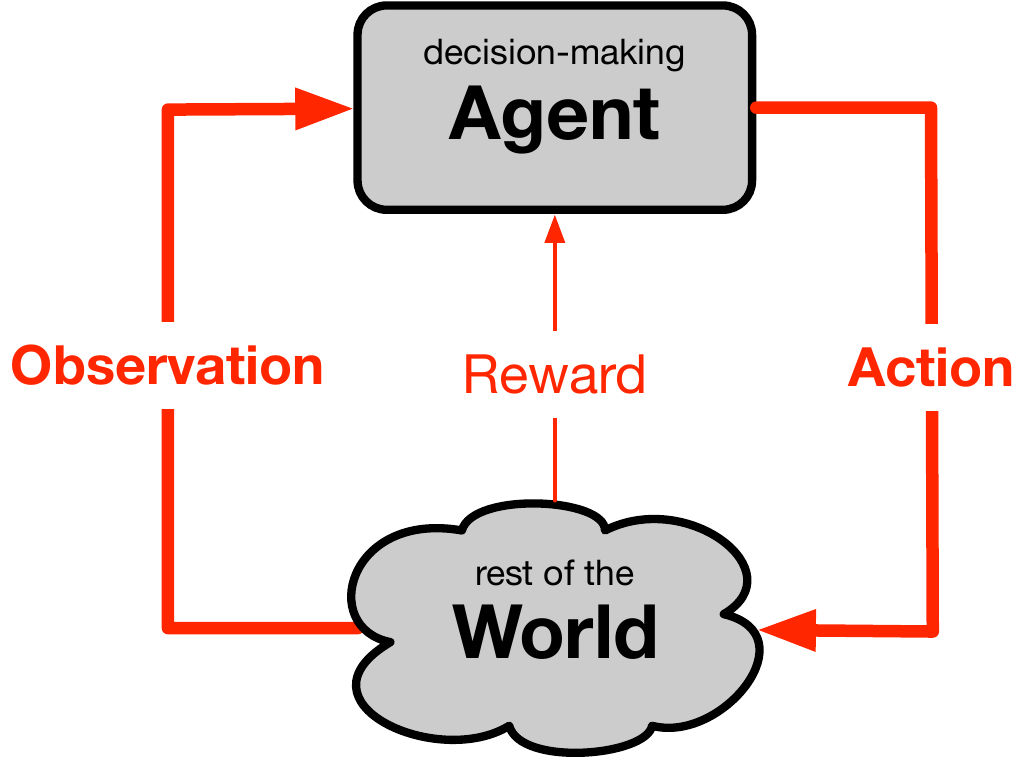}
\end{wrapfigure}
What then does the decision-making agent interact with? One answer is that it interacts with everything that is not the agent, which might be termed its ``environment" or ``world." Either term is good for our purposes---not strongly linked to a specific discipline---but let's go with \emph{world} for this paper just because it is simpler and shorter while also being evocative in a way not linked to any specific discipline. To complete the picture of the agent--world interaction (right) we must give names to the signals passing in each direction. Without belaboring the points, it is natural to say that the agent takes \emph{actions}, and receives ``sensations" or ``observations." Let us use \emph{observations} because it is an established term for this purpose and avoids metaphysical discussions about whether a machine can have ``sensation." In its standard usage, ``observations" means information about the state of the world that is potentially incomplete. Reward will be discussed shortly.

\section{Ground Rules}

The preceding discussion exemplifies the ground rules and the steps that I attempt to follow regarding terminology: 1) identify the discipline-independent idea that the word is meant to connote, and 2) find a commonsense word that captures that meaning without being unduly biased toward one discipline or another. And finally: 3) repeat the first two steps until consensus has been found across disciplines. 

A second kind of ground rule that I follow is not about terminology, but about content. As we attempt to develop a common model of decision making, what aspects should we include and which exclude? The rule I attempt to follow is to include the intersection of the fields rather than the union. That is, in order for a aspect to be included it is not enough for it arise in just one of the fields. There must be at least be an argument that it is pertinent to many if not all of them. The aspects of the common model must be general to all decision making over time to achieve a goal.

Moreover, there should be nothing in the common model that is specific to our world---for example, nothing about vision, objects, 3D space, other agents, or language.
Simple examples of what we exclude are all the things that make people special and distinct from other animals, or all the special knowledge that animals have built into them by evolution to fit to their ecological niches. 
These things are vitally important topics in anthropology and ethology, and genuinely improve our understanding of naturally intelligent systems, but have no place in the common model.
Similarly, we exclude all the domain knowledge that is build into artificial intelligent systems by their human designers in order to make a more useful application that requires less extensive training.
All of these things are important within their discipline separately, but are not relevant to a common model intended to apply broadly across disciplines.

A common model of decision making could have utility beyond facilitating cross-disciplinary interaction.
It is relatively easy to see the benefits within each each discipline, when they occur, of a common model because the existing disciplines and their value are already established. Understanding natural systems has clear scientific value. Creating more useful engineered artifacts has clear utilitarian value. But there is also scientific value in understanding the process of intelligent decision making independent of its relationship to natural decision making, and independent of the practical utility of its artifacts. I think so. It is not a currently established science, but one day perhaps there will be established a science of decision making independent of biology and of its engineering applications.

\section{Additive Rewards}

We turn now to the decision-making agent's goal.
Today most disciplines formulate the agent's goal in terms of a scalar signal generated outside the agent's direct control, and thus we place its generation, formally, in the world.
In the general case this signal arrives on every time step and the goal is to maximize its sum. Such \emph{additive rewards} may be used to formulate the goal as a delay-discounted sum, as a sum over a finite horizon, or in terms of average reward per time step.
Many other names have been used for reward, such as ``payoff," ``gain," ``utility," and, when minimized rather than maximized, ``costs."
Formulation in terms of ``costs" and minimization is formally equivalent if the costs are allowed to be negative, but just a little more complicated.
A simpler but still popular notion of goal is as a state of the world to be reached.
The goal-state formulation is sometimes adequate, but less general than additive rewards. For example, it cannot handle goals of maintenance, or specify how time-to-goal and uncertainty are traded off, 
all of which are easily handled with additive frameworks.

Additive rewards have a long inter-disciplinary history.
In psychology ``reward" was used primarily for external objects or events that were pleasing to the animal, and even if that pleasing-ness was derived from an association of the object with something that was rewarding in a more basic way---a ``primary reinforcer." Today's usage of ``reward" in operations research, economics, and artificial intelligence is restricted to that more primary signal, and is also more explicitly a received signal rather than being tied to external objects or events.
That usage appears to have been established with the development of Markov decision processes within optimal control and operations research in the 1960s. 
It is now standard in an impressively wide range of disciplines, including economics, 
reinforcement learning, 
neuroscience, 
psychology, 
operations research, 
and multiple subfields of artificial intelligence. 

\section{Standard Components of the Decision-Making Agent}

We turn now to the internal structure of the agent. This may be the topic on which it most difficult to obtain multi-disciplinary consensus. I have opted to include in the agent only the most essential elements for which there is widespread (albeit not universal) agreement within and across disciplines, and to describe them only in general terms. Even so, I hope that my selections and descriptions may be useful for communication across disciplines and that even those who disagree with them will see my choices as representative of a plurality of researchers in the field. Even if a view turns out to be wrong, it can still be a contribution to make it clear.

\begin{wrapfigure}{R}{0.5\textwidth}
\includegraphics[width=0.5\textwidth]{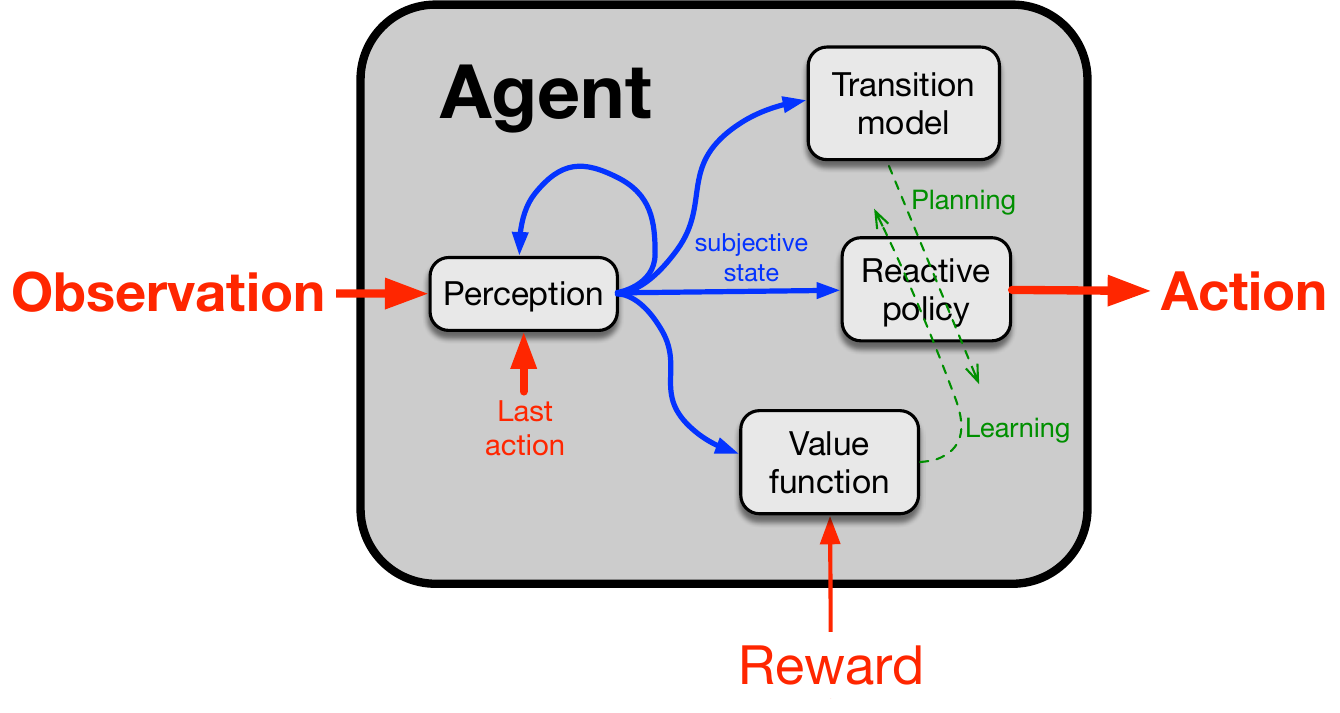}
\end{wrapfigure}
The proposed common model of the internal structure of the agent has four principal components---\emph{perception}, \emph{reactive policy}, \emph{value function}, and \emph{transition model}---shown in the figure below as boxes interconnected by a central signal, the ``subjective state." 
Although the four components are each common to many disciplines, it is rare for all to be present in fully developed form within a single agent, and of course specific agents may have additional components beyond these four.
Let us consider each component of the common model in turn.

The \emph{perception} component processes the stream of observations and actions to produce the \emph{subjective state}, a summary of the agent--world interaction so far that is useful for selecting action (the reactive policy), for predicting future reward (the value function), and for predicting future subjective states (the transition model). The state is subjective in that it is relative to the agent's observations and actions and may not correspond to the actual internal workings of the world. 
Often the construction of subjective state is a fixed pre-processing step, in which case the agent is assumed to receive the subjective state directly as an observation. For example, in Atari game playing, the subjective state may be the last four video frames together with the most recent action. 
In Bayesian approaches the subjective state \emph{does} have a  relationship to the internal workings of the world: the subjective state is intended to approximate the probability distribution over the latent states that the world uses internally (as in Partially Observable Markov Decision Processes, or Kalman filters). 
In predictive state methods (e.g., Littman et al.\ 2002) the subjective state is a set of predictions. 
In deep learning, the subjective state is typically the transient activity of a recurrent artificial neural network (e.g., Hochreiter \& Schmidhuber, 1997). 
In control theory, the computations of the perception component are often referred to as `state identification," or ``state estimation." 
In general (and in many of these examples) the perception component should have a \emph{recursive form} allowing the subjective state to be computed efficiently from the preceding subjective state, the most recent observation, and the most recent action, without revisiting the lengthy history of prior observations and actions. Processing in the perception component needs to be \emph{fast}, that is, completed well within the time interval between successive time steps of the agent--world interaction.

The perception component is not just about short-term memory, but also about representation. For example, it includes any domain-dependent feature construction processes such as are commonly used in classical pattern recognition systems. More generally, 
when a person observes a complex visual image and re-represents it in terms of objects and relationships, or observes a chess position and then re-represents it in terms of threats and pawn structure---these things too are done in the perception component. The perception component converts the stream of observation--action events into a summary representation that is relevant to the current moment. Such examples suggest the ways in which the term ``perception" is appropriate here: one dictionary definition of perception is ``the neurophysiological processes, including memory, by which an organism becomes aware of and interprets external stimuli." Generalizing that beyond biology, perception is an entirely appropriate name for the recursive updating of the agent's state representation.

The \emph{reactive policy} component of the common model maps the subjective state to an action. Like perception, the reactive policy must be fast; the speed of perception and the reactive policy together determine the overall reaction time of the agent. Sometimes perception and the reactive policy are treated together, as in end-to-end learning, 
but still both functions are generally identifiable. Separating overall action generation into these two parts---perception and policy---is common in many disciplines.
In engineering it is common to assume that perception is given, not learned, and not even part of the agent. Engineering clearly has the idea of a reactive policy, usually computed or derived analytically. Artificial intelligence systems more often assume that substantial processing is possible before acting (e.g., chess-playing programs). Classical psychology has its Stimulus-Response associations. In psychology it is common to view perception as some that supports but precedes action and can be studied independently of its effects on particular actions.

The \emph{value function} component of the common model maps the subjective state (or state--action pair) to a scalar assessment of its desirability, operationally defined as the expected cumulative reward that follows it. This assessment is fast and independent of its justification (like an intuition) but may be based on long experience (or even on expert design or generations of evolution) or from extensive computations that are effectively stored or cached. In either way, the assessments are such that can be called up quickly to support processes that alter the reactive policy. 

Value functions have a very broad multi-disciplinary history. They were first extensively developed, in optimal control, operations research, and dynamic programming, as ``cost-to-go" functions satisfying Bellman equations and, in continuous time, Hamilton-Jacobi-Bellman equations. In economics they are called ``utility functions" and appeared initially as inputs to economic calculations capturing consumer preferences. In psychology, they are related to old ideas of secondary reinforcers and to newer ideas of reward prediction. The terminology of value functions came initially from dynamic programming and was then taken up within reinforcement learning, where value functions are used extensively as critical components of the theory and of almost all learning methods. Artificial intelligence and operations research also have a long history of search methods that operate on approximate cost-to-go functions. 
In neuroscience, the error in the value function, or ``reward-prediction error," has been hypothesized as an interpretation of the phasic signal of the neurotransmitter dopamine (Schultz, Dayan \& Montague 1997). The theory has explained a wide range of puzzling empirical results and has been extremely influential (e.g., see Glimcher 2011).

The fourth and last component of the common model of the agent, the \emph{transition model}, takes in states and predicts what next states would result if various actions were taken. The transition model could be termed a ``world model," 
but this would overstate its role, as much of what constitutes a model of the world is actually in the perception component and its state representations. The transition model is used to simulate the effects of various actions and, with the help of the value function, evaluate the possible outcomes and change the reactive policy to favor actions with predicted good outcomes and disfavor actions with predicted poor outcomes. When this process is done without actually taking the actions, possibly even without visiting the states, then it is appropriate to call it ``planning," or even ``reason." 

Transition models play important roles in many disciplines. 
In psychology, internal models of the world such as provided by the transition model together with perception have been prominent models of thought since the work of Kenneth Craik (1943) and Edward Tolman (1948). Tolman showed that the results of many rat experiments could be interpreted as the rat forming and using a ``cognitive map" (i.e., transition model) of its world. In neuroscience, the hippocampus is now widely seen as providing such a map, and theorists including Karl Friston and Jeff Hawkins have extensively developed brain theories based on this idea. More recently, in psychology, Daniel Kahneman (2011) has proposed the idea of two mental systems, System 1 and System 2. The first three components of the common model of the agent fit squarely in System 1, and model-based prediction and planning are naturally interpretted as important parts of System 2.
In control theory and operations research it has always been common to  use transition models of many forms, 
including differential equation models, difference equations, and Markov models. Models of the world are a common assumption in classical artificial intelligence, such as in single-agent and game-tree search. 
In reinforcement learning, model-based approaches in which the model is learned have long been proposed 
and are now beginning to be effective in large applications (e.g., Schrittwieser et al.\ 2020).
In modern deep learning, prominent researchers such as Yoshua Bengio, Yann LeCun, and J\"urgen Schmidhuber all place predictive models of the world at the center of their theories of the mind.

\vspace*{-3pt}
\section{Limitations and Assessment}
\vspace*{-3pt}

This has been a short treatment of the quest for a common model of the intelligent agent. All the points briefly made here deserve elaboration and a deeper treatment of the history. Nevertheless, the main points seem clear. We have presented a prominent candidate for the common model. Its external interface---in terms of agent, world, action, observation, and reward---is general, natural, and widely assumed in both natural sciences and engineering. The four internal components of the agent also each  have a long and broad multi-disciplinary tradition.

The proposed common model could be criticized because of what it leaves out. For example, there is no explicit role in it for predictions of observations other than reward, and there is no treatment of exploration, curiosity, or intrinsic motivation. And all of the the four components must involve learning, but here we have described learning only in the reactive policy, and that only in general terms. Many readers are no doubt disappointed that the common model fails to include their favorite feature, whose importance is underappreciated. I, for example, feel that auxiliary sub-tasks posed by the agent for itself (as in Sutton et al.\ 2022) are an important and under-appreciated means by which the agent developes abstract cognitive structure. Yet, precisely because auxiliary subtasks are not widely appreciated, they should not appear in the common model of the agent; they are not sufficiently agreed upon across disciplines.

The ambition of the common model of the intelligent agent proposed here is not to be the latest and best agent, but to be a point of departure. It seeks to be a straightforward design that is well and broadly understood in many disciplines. Whenever a researcher introduces a novel agent design, the common model is meant to be useful as a standard that can be pointed to in explaining how the new design differs from or extends the common model. 

\vspace*{-3pt}
\section*{References}
\vspace*{-3pt}

\parskip=0pt
\parindent=0pt
\def\hangin{\hangindent=0.2in}
\def\bibitem[#1]#2{\hangin}
\small

\bibitem[Craik, 1943]{Craik}
Craik, K. J.~W. (1943).
\newblock {\em The Nature of Explanation}.
\newblock Cambridge University Press, Cambridge.

\hangin
Glimcher, P. W. (2011). Understanding dopamine and reinforcement learning: The dopamine reward prediction error hypothesis. \emph{Proceedings of the National Academy of Sciences, 108}(Supplement 3), 15647--15654.

\hangin
Hochreiter, S., Schmidhuber, J. (1997). Long short-term memory. \emph{Neural computation, 9}(8), 1735--1780.

\hangin
Kahneman, D. (2011). \emph{Thinking, Fast and Slow}. Macmillan.

\hangin
Littman, M. L., Sutton, R. S., Singh, S. (2002). Predictive representations of state. In \emph{Advances in Neural Information Processing Systems 14}, pp.~1555--1561. MIT Press, Cambridge, MA.

\hangin
Schrittwieser, J., Antonoglou, I., Hubert, T., Simonyan, K., Sifre, L., Schmitt, S., ... Silver, D. (2020). Mastering atari, go, chess and shogi by planning with a learned model. \emph{Nature, 588}(7839), 604--609.

\hangin
Schultz, W., Dayan, P., Montague, P.~R. (1997).
\newblock A neural substrate of prediction and reward.
\newblock {\em Science, 275}(5306):1593--1598.

\hangin
Sutton, R. S., Machado, M. C., Holland, G. Z., Timbers, D. S. F., Tanner, B., White, A. (2022). Reward-respecting subtasks for model-based reinforcement learning. arXiv:2202.03466.

\hangin
Tolman, E. C. (1948). Cognitive maps in rats and men. \emph{Psychological review, 55}(4), 189.

\end{document}